\documentclass[12pt]{article}

\usepackage[left=0.5in, top=0.75in, right=0.5in, bottom=0.75in]{geometry} 

\usepackage[T1]{fontenc}
\usepackage[osf]{mathpazo}

\usepackage{amsmath,amssymb,amsfonts}
\usepackage{algorithm,algpseudocode}
\usepackage{graphicx}
\usepackage{subfigure}
\usepackage{textcomp}
\usepackage{xcolor}
\usepackage{url}

\title{Emergent Braitenberg-style Behaviours for Navigating the ViZDoom `My Way Home' Labyrinth}

\author{Caleidgh Bayer \and
Robert J. Smith \and
Malcolm I. Heywood}
\date{Faculty of Computer Science, Dalhousie University\\Halifax, NS. Canada}

\begin{document}

\maketitle

\begin{abstract}
The navigation of complex labyrinths with tens of rooms under visual partially observable state is typically addressed using recurrent deep reinforcement learning architectures. In this work, we show that navigation can be achieved through the emergent evolution of a simple Braitentberg-style heuristic that structures the interaction between agent and labyrinth, i.e. complex behaviour from simple heuristics. To do so, the approach of tangled program graphs is assumed in which programs cooperatively coevolve to develop a modular indexing scheme that only employs 0.8\% of the state space. We attribute this simplicity to several biases implicit in the representation, such as the use of pixel indexing as opposed to deploying a convolutional kernel or image processing operators.
\end{abstract}


\section{Introduction}
Autonomous agents are considered to be \emph{emergent} when decision making appears at different `scales’. Thus, behavioural properties might emerge at \emph{slower} scales than that of the original sensor interactions \cite{gershenson23}. Likewise, the slower scale is conducive to decision making at a more abstract/higher-level than the original atomic sensor representation of state. In addition, complexity potentially results when novel information/behaviours result from the interaction between different \emph{components} of a representation \cite{gershenson23}.

In the case of a single decision making agent, emergence implies that different components interact to produce behaviours that are not possible from the individual components alone. An early example of this were Braitenberg’s reactive automatons (or vehicles) \cite{braitenberg84}. In their simplest form a Braitenberg vehicle consists of one or more sensors that enforce a reactive mechanism to define a corresponding agent behaviour. Thus, sensing light might direct the agent to move towards the light, ultimately reaching the light source (excitatory sensor feedback alone). More interesting behaviours result when oscillation takes place between excitatory and inhibitory sensory feedback (agent cycles between approaching and evading the light source). More recently, Braitenberg vehicles have been used to model behavioural properties of multiple biological organisms, e.g. fruit fly odor based foraging, bat obstacle avoidance and phototaxis in lizards \cite{shaikh20}. The robotics community has also made use of Braitenberg vehicles, given specific sensor modalities, e.g. olfactory \cite{chen19} or auditory \cite{huang99}. However, as the dimensionality of the state space increases (e.g. pixel based visual state) the ability to abstract suitably informative simple behaviours decreases. Instead, deep learning (DL) approaches have increasingly been adopted in which convolutional neural networks are used to discover embeddings from which decisions can be made. One limitation of such an approach is that it is more difficult to decompose the resulting behaviour into specific behavioural components. A second limitation is that deploying the resulting DL solution on platforms such as robots/ drones becomes increasingly intractable as the platform scale decreases.

In this work, we are interested in revisiting the ability to evolve Braitenberg-like solutions under a labyrinth described in terms of a partially observable high-dimensional visual state space ($160 \times 120 = 19 200$ pixels per frame). We adopt the genetic programming approach of tangled program graphs (TPG) \cite{kelly17,kelly18a}, but limit the approach to very small graphs (solutions can only consist of tens of programs \cite{bayer21}) and preclude the use of indexed memory, e.g. \cite{kelly21}. Our insight is that TPG provides the opportunity to discover policies with multiple programs that typically index a fraction of the available state space. Our hypothesis is that the resulting programs will describe very simple component trajectories that are suitably context dependent/general to collectively navigate the labyrinth. In effect, low dimensional components need to be discovered that enable agents to decompose the original high-dimensional partially observable task into components compatible with the original motivation for Braitenberg vehicles.

The environment assumed takes the form of the `My Way Home' (MWH) task from the ViZDoom environment \cite{kempka16}. Specifically, the MWH task requires an agent to navigate through up to 17 different room/corridor combinations to reach a common goal. Agents experience the task using high-dimensional state information defined in terms of (pseudo) 3-D video data under a first person perspective (i.e. partially observable) and sparse rewards. We note that such a task can be solved with various deep learning frameworks, but only if they are provided with suitable memory mechanisms. In contrast we are interested in the ability to discover structure through (a) trajectory sequences defined by the graph representation itself while (b) switching between behavioural components reactively.

Section \ref{sec:vizdoom} summarizes the properties of the MWH task and highlights some recent results. Section \ref{sec:methods} summarizes the properties of TPG and a comparator DL architecture. Specifically, we include the DL architecture to provide a baseline result for a memoryless agent on the MWH task to reaffirm the task difficulty. Section \ref{sec:results} performs the empirical study, ultimately confirming the ability of TPG to discover Braitenberg-style solutions to the task. Conclusions are drawn and future research motivated in Section \ref{sec:conc}.

\section{Background to the My Way Home labyrinth}\label{sec:vizdoom}

\begin{figure}
\begin{center}
\includegraphics[width=8.5cm]{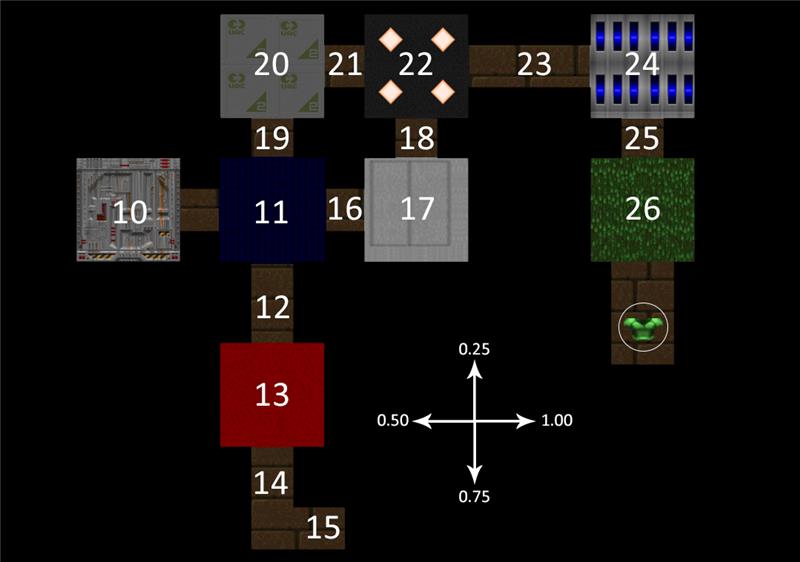}
\caption{Map of the `My Way Home' labyrinth. Room numbers are added for identification purposes alone}\label{fig:myh_map}
\end{center}
\end{figure}

The MWH task scenario\footnote{All ViZDoom scenarios defined at \url{https://github.com/Farama-Foundation/ViZDoom/blob/master/scenarios/README.md}} defines a set of 8 rooms and 10 corridors from which an RL agent has to navigate its way to a room containing a green vest (indicated with a circle in Figure \ref{fig:myh_map}). The underlying objective is to learn a policy for successfully navigating from any location to the corridor with the green vest. Agents do not perceive state through the `birds eye' view of Figure \ref{fig:myh_map}, but through a visual first person perspective defined relative to the orientation of the agent (i.e. partially observable state). The agent may move through the labyrinth using three actions: move forward, turn left, turn right. At each time step, \emph{t}, the agent receives a reward, $r_t = -0.0001$ with a (max. episode length) of 2,100 time steps. An episode may end earlier if the agent finds the vest, i.e. a reward of $r_t = 1.0$ is received, otherwise an episode ends with a reward of $r_{t=2,100} = 0$. In summary, finding a path to the vest is better than not finding a path, and shorter paths are better than longer ones. The game engine spawns an agent in a randomly chosen location\footnote{Spawn location cannot be the corridor with the vest and must be within a legal part of the labyrinth, i.e. the numbered locations in Figure \ref{fig:myh_map}.} with uniform probability, facing a random direction. The rooms have particular colour/textures incorporated into the walls/floors whereas connecting corridors assume common visual properties.

Recent research under the ViZDoom MWH task includes the following highlights. A survey of deep RL for exploration in complex environments \cite{hao23} identified the use of intrinsic motivation as being particularly effective under the MWH navigation task. Formulations for intrinsic motivation are based on episodic memory that receives a novelty bonus \cite{savinov18}. Gated Recurrent Units (GRU) represent a reoccurring memory architecture that often appears as a building block for Deep RL. For example, the Advantage Actor Critic formulation benchmarked on MWH assumes GRU for internal state \cite{beeching20}, likewise PPO with GRU. However, neither solve the task with the GRU disabled. In short, solutions to partially observable navigation tasks such as MWH typically assume the use of (a) a mechanism for embedding full state information (convolution operator) in a lower dimensional representation (b) some form of memory. In the case of TPG, \cite{smith18} evolved solutions to a total of 10 different task scenarios \emph{simultaneously} without memory. As such the resulting solution indexed just over half of the state space ($\approx$ 11,000 pixels) and consisted of over 17 thousand instructions. The hypothesis of our work is that this complexity is not necessary. Instead, we believe that partially observable high-dimensional navigation tasks can be reduced to simple Braitenberg reactive behaviours.

\section{Methods}\label{sec:methods}

\subsection{Deep Q-Network}\label{sec:dqn}

DQN demonstrated that the method of temporal differences (Q-learning specifically) could scale to state spaces described by (quantized and stacked) video sequences \cite{mnih15}. Q-learning is an off-policy gradient approach for addressing the issue of delayed rewards that characterizes reinforcement learning tasks \cite{watkins92}. Some key elements of DQN include the use of a deep learning architecture to support the `hierarchical sensory processing' of the original visual state space across many tasks. Hence DQN does not rely on feature engineering for each task. DQN also introduced `experience replay' and `iterative action-value updates' in order to reduce correlation with input state and target values respectively \cite{mnih15}. To do so, a planning approach\footnote{Planning in RL implies that a policy can be updated without further references made to the environment, therefore more query efficient. Naturally, planning and interaction with the task have to be interleaved in order to expose an agent to new aspects of the task.} is adopted in which a history of the last million state-action-reward tuples are stochastically sampled in order to provide the basis for mini-batch weight optimization. Moreover, DQN provided the original demonstration for visual reinforcement learning in both the Atari 2600 and ViZDoom environments (e.g. \cite{mnih15} and \cite{kempka16} respectively). The history is not considered memory because tuples are limited to state--action--rewards. However, the aforementioned stacking does embed each state observation with a sequence of 5 consecutive frames. DQN therefore represents a known baseline agent for visual reinforcement learning tasks to qualify the impact of partial observability in the MWH task.

\subsection{Tangled Program Graphs}\label{sec:tpg}

TPG approaches the issue of scaling to visual reinforcement learning tasks through a combination of sparse state sampling, emergent modularity, and averaged sample returns \cite{kelly17,kelly18a}. Specifically, as a genetic programming approach, individuals rarely sample the state space in its entirety, i.e. the agent has to determine what pixels from the state space need indexing. Emergent modularity refers to the ability of TPG to incrementally compose a solution from multiple programs. Teams of graphs of programs ultimately result where different parts of the graph are visited conditionally depending on the (task) state. Credit assignment assumes a process by which (1) reward is averaged over an episode, i.e. once a terminating state is encountered (max. number of interactions with the task or a goal/failure condition is encountered) and (2) multiple candidate solutions are maintained and ranked relative to each other. Reactive formulations of TPG have previously been benchmarked under both Atari 2600 and ViZDoom environments \cite{kelly17,smith18,kelly18b}. For completeness, we summarize key properties of TPG below and refer interested readers to \cite{kelly18a,kelly18b,kelly21} for tutorial treatments.

\subsubsection{Programs} In this work, we assume that programs define a sequence of instructions parameterizing an imperative programming language, i.e. linear genetic programming \cite{brameier07}. Each instruction is defined in terms of a tuple,

\begin{equation}%
\langle \mbox{\ttfamily mode, target, op, source} \lor \mbox{\ttfamily input} \rangle
\end{equation}

\noindent Thus, an instruction manipulates the values in a set of general purpose registers ({\ttfamily R[$\cdot$]}) using the following general form, 

\begin{equation}%
\mbox{\ttfamily R[target]} \leftarrow \mbox{\ttfamily R[target]} \langle \mbox{\ttfamily op} \rangle \mbox{\ttfamily R[source]} \lor \: \vec{s} \mbox{\ttfamily [input]} 
\end{equation}

\noindent where {\ttfamily mode} $\in [0, 1]$ is used to select between {\ttfamily source} and {\ttfamily input} indexing. This also implies that $0 \leq$ {\ttfamily target, source} $ < MaxReg$ and the range for {\ttfamily input} matches the number of pixels in the state space, $\vec{s}$. Finally, {\ttfamily op} denotes the instruction set which in this case takes the form of a set of arithmetic operations $\{+. -. \div, \times\}$.

\subsubsection{Learner ($L_i$)} A tuple defines a learner comprising of a context program, $p_i$, and either an action program, $a_i$, or a pointer, $E_i$. Learners appear in a learner population, $\mathcal{L}$, and are initialized with null pointers. Context programs define which state attributes to index and therefore the state under which an action program will be called upon to define an action. Action programs define a vector of action values \cite{bayer21,amaral22}. The same action and context program can appear in multiple learners, i.e. $L_i$ is unique if the sampling of action and context program is unique.

\subsubsection{Ensemble ($E_j$)} Each TPG agent is defined as a unique ensemble of 2 to 4 learners from the learner population \cite{bayer21}. A learner may appear in multiple ensembles and ensembles are defined in a separate ensemble population, $\mathcal{E}$. The requirements for defining a legal ensemble are that: a) the complement of learners is unique, b) the min (2) and max (4) ensemble limits are not passed, and; c) there are at least two different action programs appearing in the ensemble.

\subsubsection{Ensemble evaluation}\label{sec:eval_team} Evaluating an ensemble begins with a state observation from the task, $\vec{s}(t)$, at time step \emph{t}, i.e. current state. In the case of ViZDoom state is a matrix of pixels, thus an array of indexed attribute values {\ttfamily $\vec{s}$[a]}. Given a specific ensemble, $E_j$, all the context programs for learners appearing in this ensemble are then evaluated on the current state. The respective values for register {\ttfamily R[0]} are then compared to find the winning learner

\begin{equation}
i^* = \arg \max_{L_{p_i} \in E_j} (L_{p_i}.\mbox{{\ttfamily R[0]}})
\end{equation}

\noindent where $L_{p_i}.${\ttfamily R[0]} is the value returned by register {\ttfamily R[0]} from the \emph{i}-th learner's context program $p_i$.

The action program for winning learner $i^*$ is then executed on the same state, $\vec{s}(t)$, resulting in the registers for the action program taking state-specific values. Under the ViZDoom task scenarios, actions take the form of a vector of actions from a set of task-specific atomic actions, $\mathcal{A}$. Thus, an action program has at least $|\mathcal{A}|$ registers. Action selection, $a_t$, takes the form of selecting the action corresponding to the register with the highest value after program execution.

The process of winning context program identification and action selection results in a sequence of interactions with the environment of the form: $\vec{s}_t, a_t, r_{t+1}, \cdots, \vec{s}_{t+n}, a_{t+n}, r_{T_{max}}$ where \emph{r} is a scalar reward received from the task environment at time step \emph{t} and $r_{T_{max}}$ implies that either the goal state was encountered ($r_{T_{max}} = 1.0$ and $T_{max} < 2,100$) or the maximum interaction limit was reached ($r_{T_{max}} = 0.0$ and $T_{max} = 2,100$).

\subsubsection{Variation} A `breeder' model of evolution is assumed in which all TPG agents are evaluated from a set of start states, resulting in the average performance across all training episodes per agent. Agents are then ranked, and the worst \emph{Gap}\% of agents are deleted from the ensemble population. Any members from the learner population not associated with surviving ensembles are then also deleted. Variation begins by selecting \emph{Gap}\% members from the ensemble population as parents with uniform probability. Denote this set the \emph{offspring set} where the offspring are initially clones of the original parent. Variation operators perform crossover between pairs of offspring and mutates the resulting offspring ensemble learner complement using add/delete operations. 

Variation extends to the learners comprising the offspring pool by stochastically selecting a learner, cloning it and introducing variation to the programs associated with the cloned learner (multiple forms of mutation). One of the unique aspects of this process is action mutation (Algorithm \ref{alg:m_action}). 

\begin{algorithm}
\begin{algorithmic}[1]
\If {$rand > P_{mn}$} \label{alg:m_action}
	\State {no mutation}
\Else
	\If {$rand > P_{action}$} \label{alg:switch}
		\State {$a_i \leftarrow$ Mutate($a_i$)} \label{tpg:a_term}
		\State {$E_i = \emptyset$}
	\Else
		\State {$E_i \leftarrow$ Choose($\mathcal{E}$)} \label{tpg:r_team}
		\State {Disable($a_i$)}
	\EndIf
\EndIf
\end{algorithmic}
\caption{Mutating the action type. \emph{Choose} function returns a pointer to an ensemble from the parent pool. \emph{Mutate} applies mutation operations to the action program}\label{alg:m_action}
\end{algorithm}

Given an action program, $a_i$, the first question is whether to modify it or not (Step \ref{alg:m_action}). If so there are then two choices: 1) modify the action program (Step \ref{tpg:a_term}) or 2) defer decision-making to another ensemble (Step \ref{tpg:r_team}). This later process disables the action program and replaces it with a pointer, $E_i$, to another ensemble (chosen with uniform p.d.f.). Such a process provides TPG with the ability to incrementally construct graphs of ensembles.

\subsubsection{Graph evaluation} TPG agents are those that have an in-degree of zero. Initially, all ensembles are initialized using learners with disabled ensemble pointers (all $E_i = \emptyset$), thus all members of the ensemble population have a zero in-degree. Such agents are referred to as root ensembles. Thereafter, Algorithm \ref{alg:m_action} may incrementally replace action programs with pointers to other ensembles. This means that over time the number of root ensembles will fluctuate. Evaluation always commences relative to a root ensemble. If the winning learner's `action' is a pointer to another ensemble ($E_i \neq \emptyset$) then decision-making on the current state is deferred to the ensemble identified by $E_i$. The process of ensemble evaluation (\S \ref{sec:eval_team}) then repeats until an action program declares an action.

\subsubsection{Champion agent} 
\begin{figure}
\begin{center}
\includegraphics[width=5.5cm]{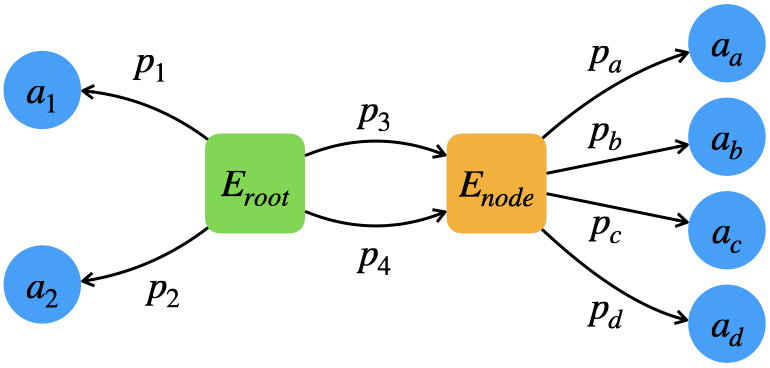}
\caption{Illustration of an example TPG champion. $E_{root}$ is the root ensemble from which evaluation always commences, arcs represent context programs ($p_i$) and leafs represent action programs ($a_i$). In this example, there are 4 learners $\in E_{root}$ of which 2 consist of context programs and a corresponding action program ($\langle p_i, a_i \rangle : i \in \{1, 2\}$) and 2 consist of context programs and pointer to the second ensemble, $E_{node}$. All 4 learners in ensemble $E_{node}$ consist of context programs and a corresponding action program ($\langle p_j, a_j \rangle : j \in \{a, \cdots, d\}$)}\label{fig:tpg_champ}
\end{center}
\end{figure}

Post training a validation step is performed to identify a champion, i.e. evaluate all root nodes given a sample of task start states. Figure \ref{fig:tpg_champ} illustrates one such champion (corresponds to the champion later used for behavioural analysis in Section \ref{sec:results}). In this case the root ensemble, $E_{root}$, consists of 4 learners, thus 4 outgoing arcs, $p_{1,...,4}$. Two arcs terminate with corresponding action programs, $a_{1, 2}$, whereas the other two arcs, $p_{1, 2}$, identify the same single (non-root) ensemble, $E_{node}$. This implies that there are two distinct contexts, $p_3$ and $p_4$, under which $E_{node}$ is deployed. This ensemble consists of another 4 learners, all of which consist of context--action program pairs, e.g. $\langle p_a, a_a \rangle$. Note, however, that no attempt has been made to remove `hitchhikers', i.e. learners with context programs that never win a round of bidding.

\section{Results}\label{sec:results}

\subsection{Parameterization}
TPG parameterization assumes the framework established by Bayer et al. (\cite{bayer21}) and are summarized in Table \ref{tbl:tpg_param}. In particular, the number of generations is applied in two phases, P1 and P2. During P1 ($g \leq 500$) $P_{action} = 0$, thus the action mutation of Algorithm \ref{alg:m_action} can only modify action programs and graphs are therefore limited to single nodes. During P2 ($g > 500$) $P_{action} = 0.5$ and action programs can be replaced with pointers to other ensembles (leading to graph construction). `Population size' reflects the number of teams with `Gap' percent replaced each generation. `Mutate Count' reflects the number of mutation operations applied to a program when a program is selected for modification, i.e. if mutation is selected the operator is applied this many times. `Evaluations' reflects the number of initializations per agent fitness assessment. The probability of add/delete a learner is used to modify parent teams (once cloned). Likewise, learner programs may also be modified by three types of mutation (add, delete, swap instructions) with probabilities of 0.5, 0.5, 1.0. Finally, 5 independent trails are performed and five champions identified. The ViZDoom environment is deployed with a frame size of $160 \times 120$. DQN parameterization assumes that from the ViZDoom distribution.\footnote{\url{https://github.com/Farama-Foundation/ViZDoom/blob/master/examples/python/learning_pytorch.py}} 

\begin{table}
\caption{TPG parameterization}\label{tbl:tpg_param}
\begin{center}
\begin{tabular}{|c|c||c|c|} \hline
Parameter & Value & Parameter & Value \\ \hline
Generations (P1) & 500 & Generations (P2) & 500 \\
Population Size & 120 & Gap & 50\% \\
Mutate Count & 5 & Evaluations & 5 \\
Prob. delete Learner & 0.7 & Prob. add Learner & 0.7 \\
Max. \# instructions & 128 & Max. \# Registers & 8 \\
Max. Team Size & 4 & Min. Team Size & 2 \\ \hline
\end{tabular}
\end{center}
\end{table}%

The ViZDoom environment is deployed such that spawn points are selected with uniform probability across the set of rooms and corridors (Figure \ref{fig:myh_map}). The goal location remains unchanged. Testing was then performed by sampling each room uniformly and spawning the agent 100 times at different orientations. We can then measure the degree to which agents were able to navigate their way to the vest (Figure \ref{fig:myh_map}).

\subsection{Performance Function}
In the case of both TPG and DQN, the agent receives a reward of $r_t = -0.0001$ per time step. An episode is terminated if (1) the agent encounters the vest, resulting in a reward of $r_t = 1.0$ or (2) the number of time steps, \emph{t}, reaches the limit of $2,100$, resulting in a reward of $r_t = 0.0$. TPG experiences reward episodically, thus an episodic fitness of $\sum_t r_t$. The agent is initialized 5 times and the average episodic fitness represents the agent's overal fitness. DQN experiences reward incrementally through a discounted reward that lets the agent experience incremental rewards \emph{during} the episode and therefore perform gradient decent (i.e. an off-policy Q-learning formulation of temporal differences with batch updates \S\ref{sec:dqn}). The environmental feedback and terminal conditions in both cases is exactly the same.

\subsection{Performance Evaluation}

\begin{table}
\caption{Test performance on MWH task scenario. Values reflect cumulative reward received over 100 initializations per room/corridor for each of the 17 room/corridors (Figure \ref{fig:myh_map})}\label{tbl:MWHome}
\begin{center}
\begin{tabular}{|c|c|c|c|c|} \hline
Agent & average & median & best & worst\\ \hline
DQN & -0.094 & -0.21 & 0.994 & -0.21 \\
TPG & 0.561 & 0.867 & 0.997 & -0.21 \\ \hline
\end{tabular}
\end{center}
\end{table}%

\begin{figure*}
\begin{center}
	\subfigure[DQN]{\includegraphics[width=12cm]{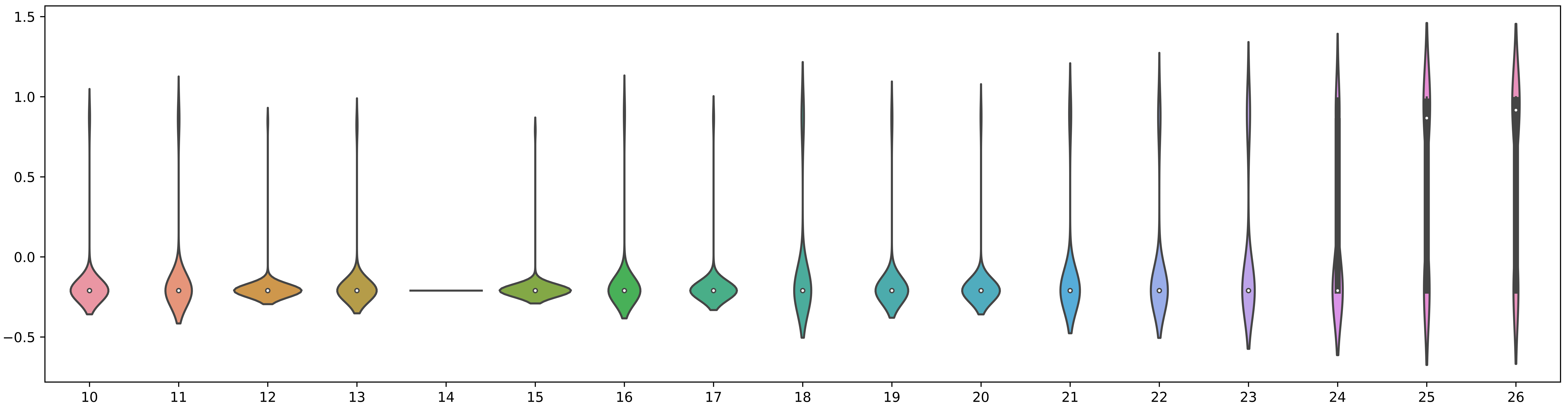}\label{fig:DQN_violin}}\qquad
	\subfigure[TPG]{\includegraphics[width=12cm]{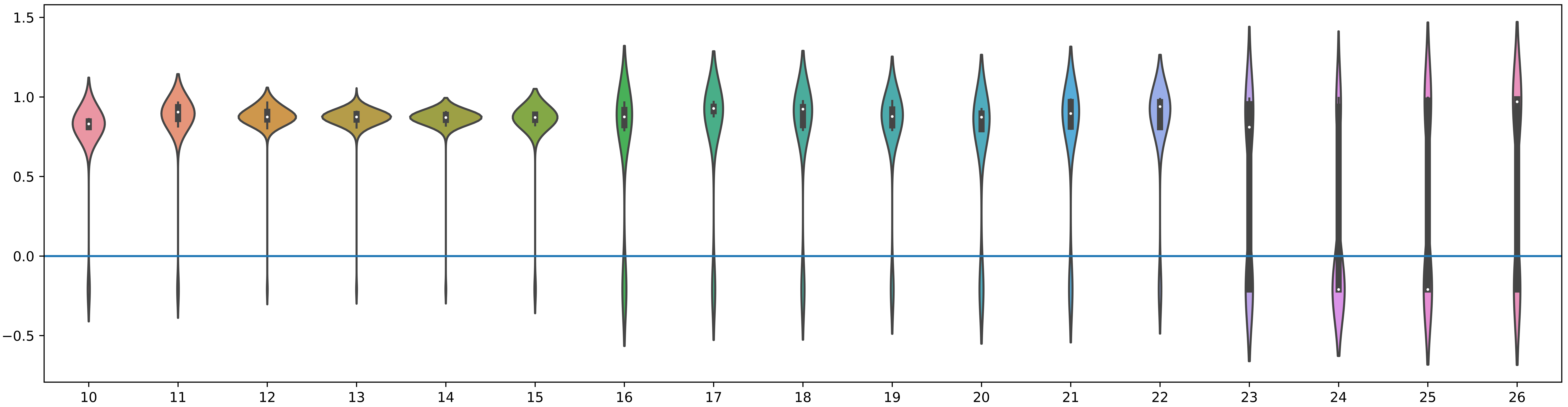}\label{fig:TPG_violin}}
\caption{Distribution of rewards under uniform test conditions. 100 spawns at random orientations per room. Room to numerical labels summarized in Figure \ref{fig:myh_map}. DQN (a) versus TPG (b). Positive values imply success (negative failure) in reaching the goal}
\end{center}
\end{figure*}

Table \ref{tbl:MWHome} summarizes the cumulative reward for champion DQN and TPG agents under test conditions. Negative scores imply that an agent failed to find a path and positive scores imply that a successful path was found. It is clear that DQN was not able to find solutions to the task whereas TPG generally was able to do so. That said, the high and low scores for both DQN and TPG are the same, indicating that room specific preferences might be present. With this in mind, a violin plot is used to summarize room specific test performance (each room/corridor was sampled 100 times with random orientation). Figure \ref{fig:DQN_violin} summarizes the distribution of DQN agent performance per room where positive (negative) results imply success (failure) in reaching the goal. The box and circle internal to each violin summarize the range of 1\textsuperscript{st}, median and 3\textsuperscript{rd} quartiles. DQN consistently navigates from rooms 26 and 25 but otherwise fails to navigate the labyrinth, i.e. DQN only navigates to the goal from the two rooms closest to the goal condition where the vest can be visually seen.

The performance of a typical TPG agent under the same test conditions is summarized in Figure \ref{fig:TPG_violin}. It is apparent that rooms 10 through 22 all have positive violin quartiles. Indeed, only spawning the TPG agent in rooms 24 and 25 results in a negative median performance. We note that particularly clean solutions appear for rooms 10 through 15, where solutions from these starting rooms involve passing through some subset of the remaining rooms (16 through 26) to reach the goal; whereas we might have anticipated the converse.\footnote{Performing a two-tailed sign test w.r.t. the null-hypothesis (equal number of positive/negative outcomes) implies TPG provides a significant improvement at critical value of 0.05.}

\subsection{Braitenberg behavioural properties of TPG champion}\label{sec:behaviour}

\begin{figure}
\begin{center}
	\subfigure[room 25]{\includegraphics[width=8cm]{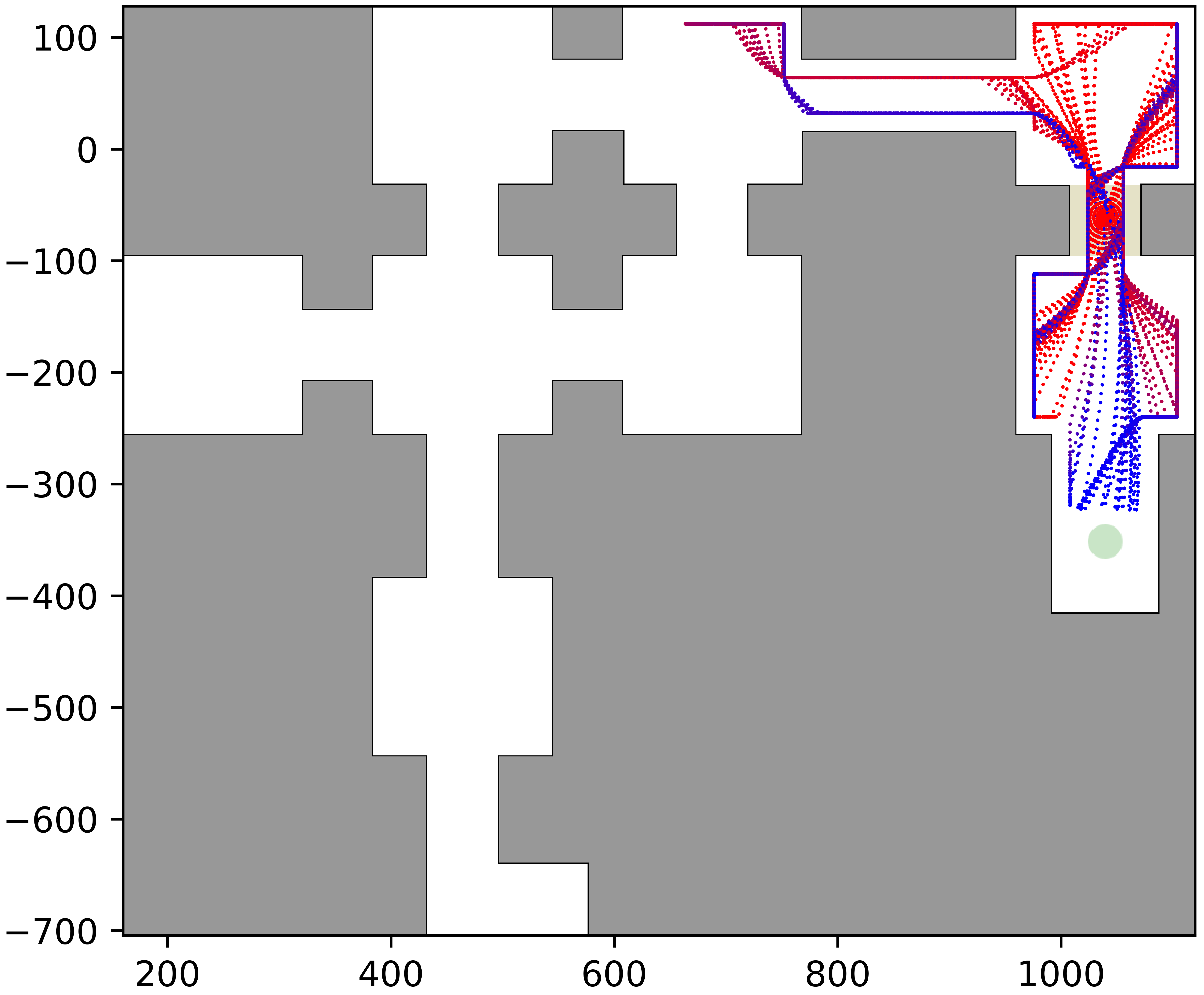}}\qquad
	\subfigure[room 22]{\includegraphics[width=8cm]{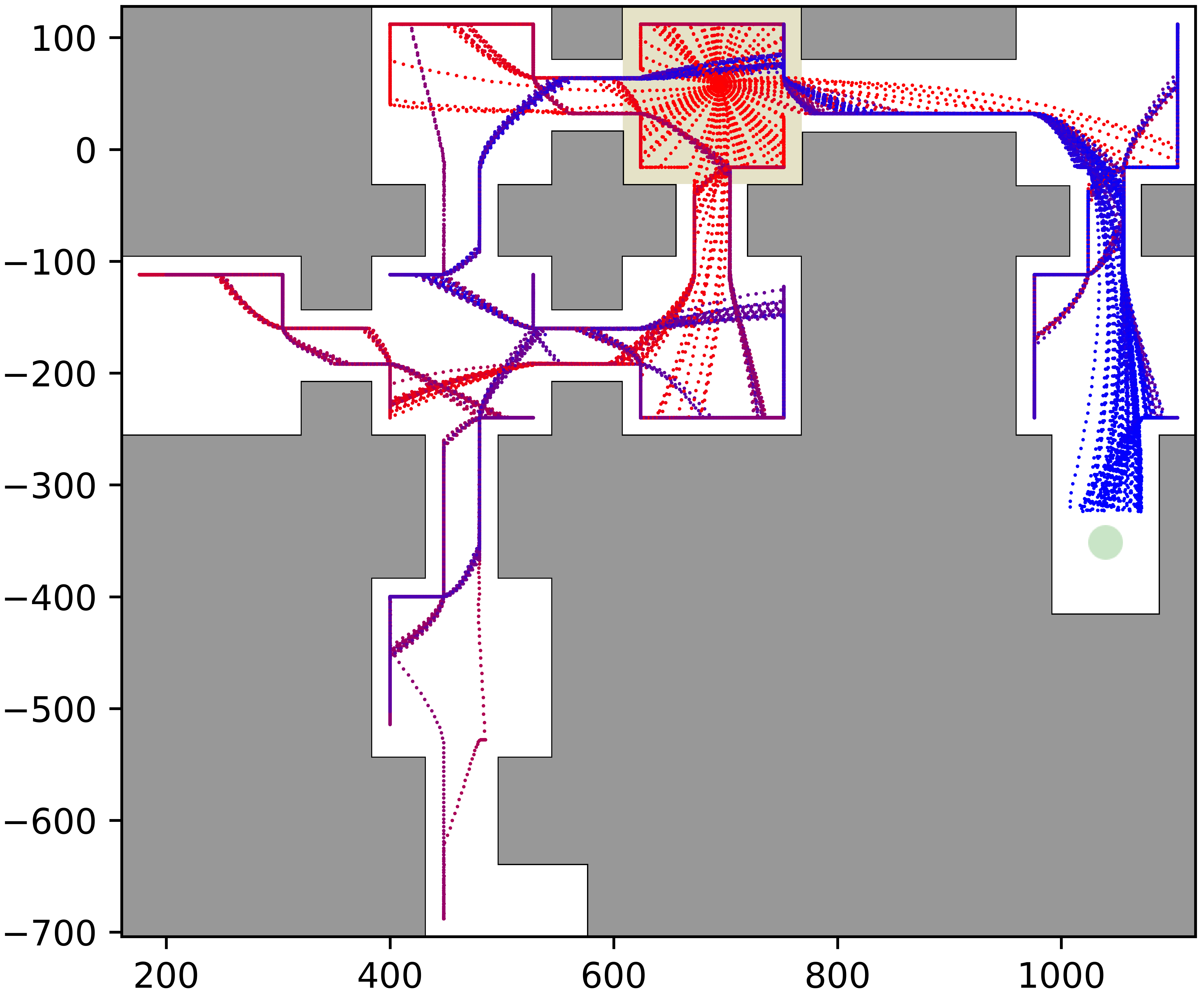}}\qquad
	\subfigure[room 15]{\includegraphics[width=8cm]{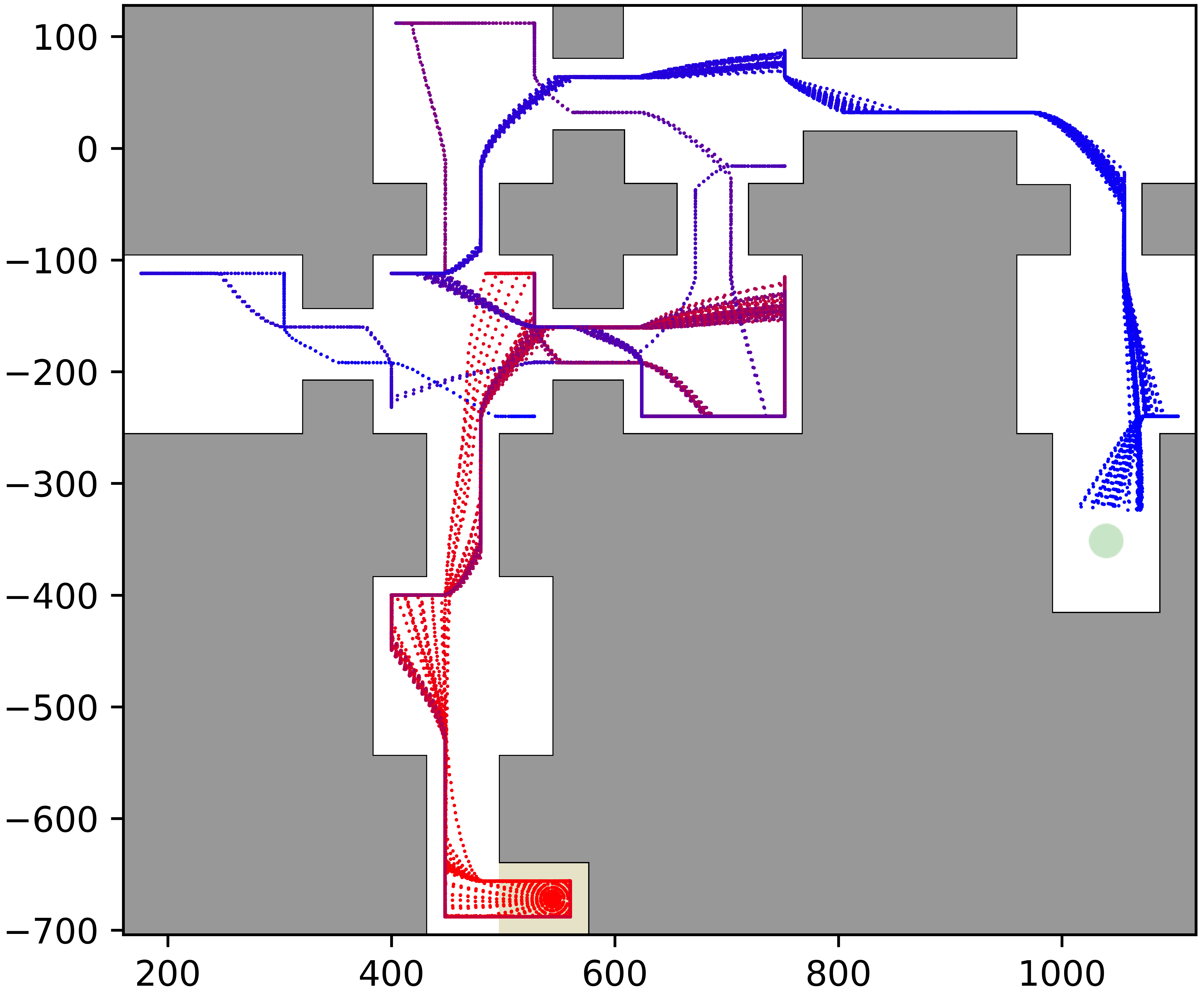}}\qquad
\caption{Example TPG solution paths for spawn points at (a) room 15 (b) room 22 and (c) room 25. Path colour transitions from red (earliest) to blue (latest) as the agent moves from spawn point to goal (or lost in the labyrinth). See Figure \ref{fig:myh_map} for a summary of labyrinth room/corridor labels}\label{fig:TPG_paths}
\end{center}
\end{figure}

Braitenberg behaviours imply that an agent's behaviour can be broken down to a set of simple heuristics that are repeatedly applied, irrespective of the starting room/corridor/orientation. With this in mind, we record the actual path taken by an agent given each of the 100 spawn points per room,\footnote{Always in the centre of the room, but with random orientation.} Figure \ref{fig:TPG_paths}. Sub-figure (a) represents the spawn point corresponding to the penultimate corridor to the goal. Depending on their initial orientation the agent will tend to use the corridor wall to exit in a northerly or southerly direction. Interestingly, after exiting the locality of the corridor the agent will then `prefer' the same side of the room that they enter next. Thus, they use the side(s) of the following room to re-orientate themselves such that they either \emph{exit} from: (1) a different corridor or (2) exit from the same corridor as they entered, but using the opposite side (from which they entered). Such behaviour is very consistent, resulting in specifically used paths for entering and exiting rooms 22 through 26.

Moving on to agent spawning from room 22 (Figure \ref{fig:TPG_paths}(b)) it is again apparent that once an exit from room 22 is taken, specific sides of a corridor are employed. This lines the agent up with a particular side for the following room. Ultimately, such a process results in the agent re-entering room 22 from the western corridor (the trajectory highlighted in blue), which then sets the agent up to exit room 22 from the eastern corridor. The consistency of such behaviour is evident from the density of the resulting (blue) paths. The majority of times the agent enters corridor `23' on a particular trajectory to line the agent up to successfully reach the goal. Moreover, the behaviour is even more apparent when room 15 represents the spawn point (Figure \ref{fig:TPG_paths}(c)). Given the consistency with which these behaviours emerge, it is apparent that a simple heuristic has emerged that succeeds in accumulating rewards for the majority of task configurations. To do so, the agent has to prioritize maximizing the average reward accumulation \emph{over the entire episode} as opposed to picking up on local reward accumulation during an episode.

In short, DQN under the My Way Home task will tend to adopt a direct line of sight behaviour. This is effective while the goal is in sight. TPG on the other hand places less emphasis on `seeing' the goal (state has to be explicitly indexed) in favour of corridor following and using rooms to re-orientate the agent. TPG therefore develops a simple reactive heuristic consistent with the overall structure of the labyrinth. We note that developing seemingly complex behaviours from simple reactive agents is a theme often associated with complex systems, e.g. Braitenberg Vehicles are capable of complex dynamic behaviours given limited state information \cite{braitenberg84,shaikh20}. However, unlike Braitenberg Vehicles, TPG discovers the reactive behaviour and state inputs necessary to solve the MWH task as opposed to having them pre-specified.

\subsection{Behaviour under no goal state}\label{sec:no_goal}
\begin{figure}
\begin{center}
	\subfigure[DQN]{\includegraphics[width=5.65cm]{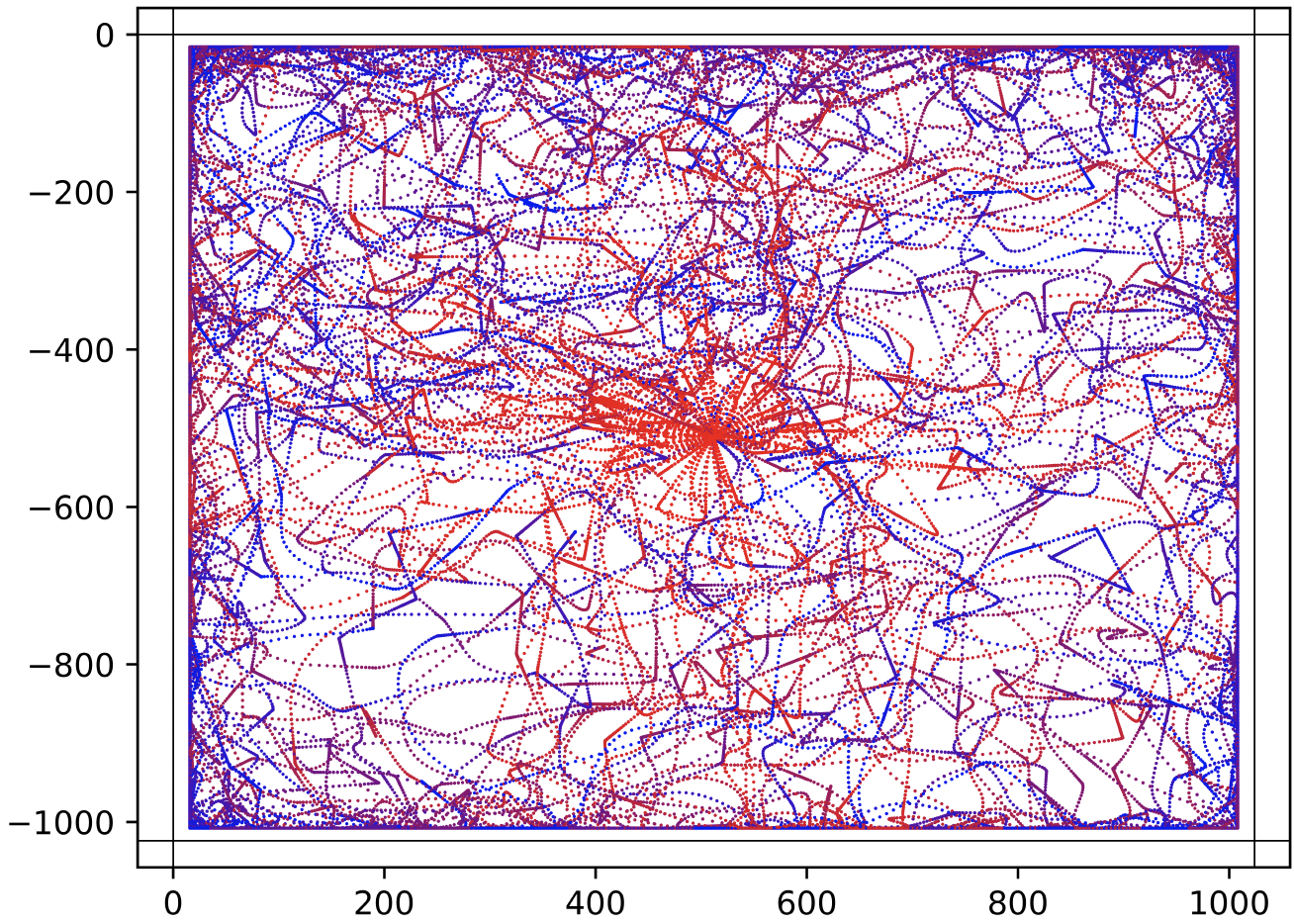}}\qquad
	\subfigure[TPG]{\includegraphics[width=5.65cm]{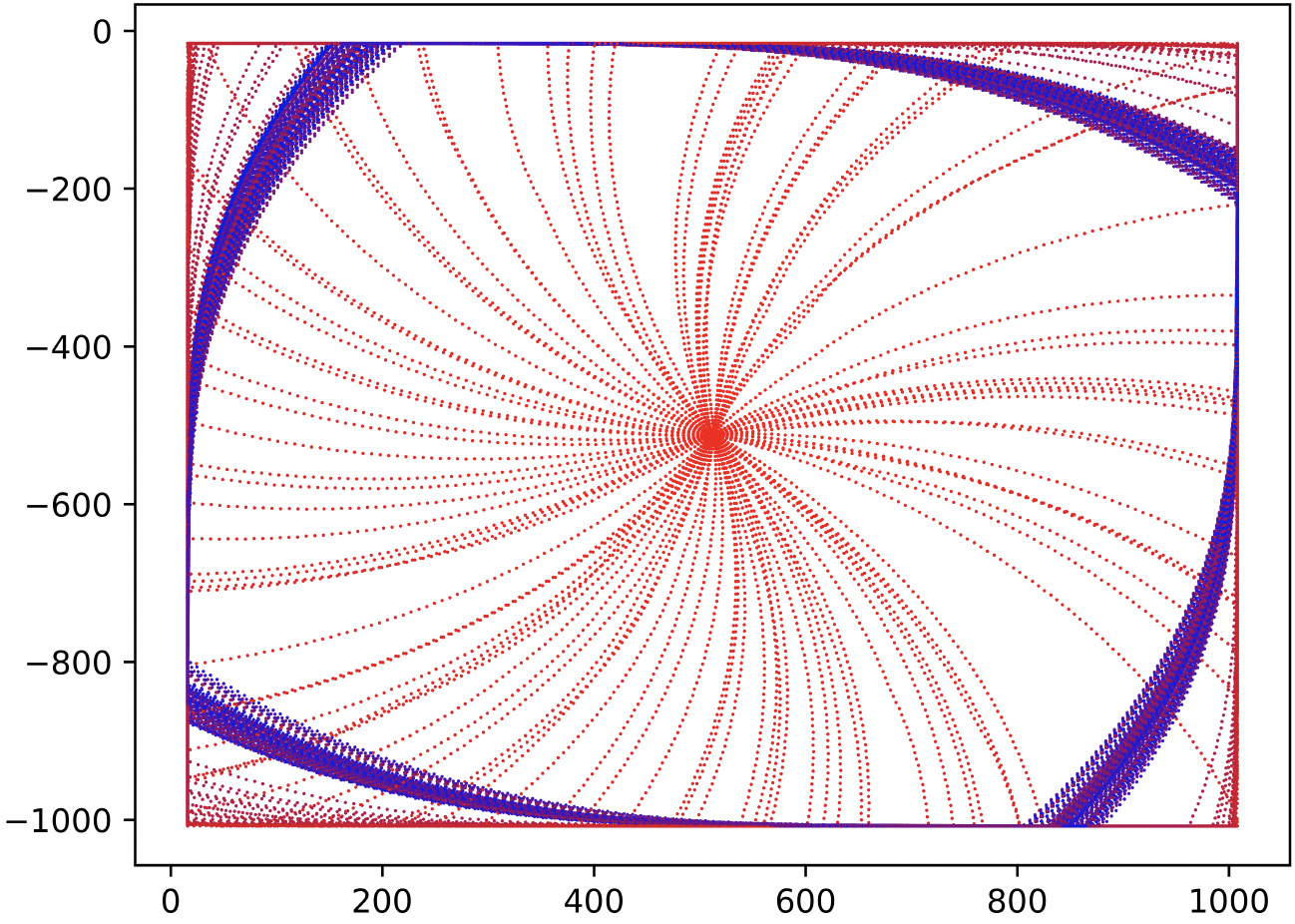}}\qquad
\caption{DQN and TPG paths in the empty room scenario. Neither DQN or TPG agents experienced this environment during training. Red (early) to blue (late) indicates the direction of time in the trajectory}\label{fig:large}
\end{center}
\end{figure}

In order to identify more clearly the properties of the TPG champion's navigation heuristic, one last experiment is performed. Post training, DQN and TPG agents are released in the centre of a single large room (as opposed to the MWH labyrinth) that assumes the texture/colour from the corridors. This room is $1000 \times 1000$ pixels, i.e. similar in size to the total footprint of the labyrinth. There is no vest, so the labyrinth goal is missing. Our motivation is that we expect the agent to repeat the same navigation heuristic, irrespective of whether the goal is present or the structure of the task. In defining a task consisting of a single room and no vest, we expect the basis for some of the navigation heuristic to be more obvious. Figure \ref{fig:large} illustrates the paths taken for a best performing DQN and TPG agent when spawned at the centre and orientated randomly.

DQN adopts a meandering policy until the wall is encountered and then essentially stays close to the wall. Without the vest, DQN is not able to apply structure to its search. TPG on the other hand adopts a slowly arcing path until the wall is encountered. It then aligns itself with the wall and makes another slow arcing path, eventually encountering the next wall with the policy repeating thereafter.

\begin{figure*}
\begin{center}
	\includegraphics[width=13cm]{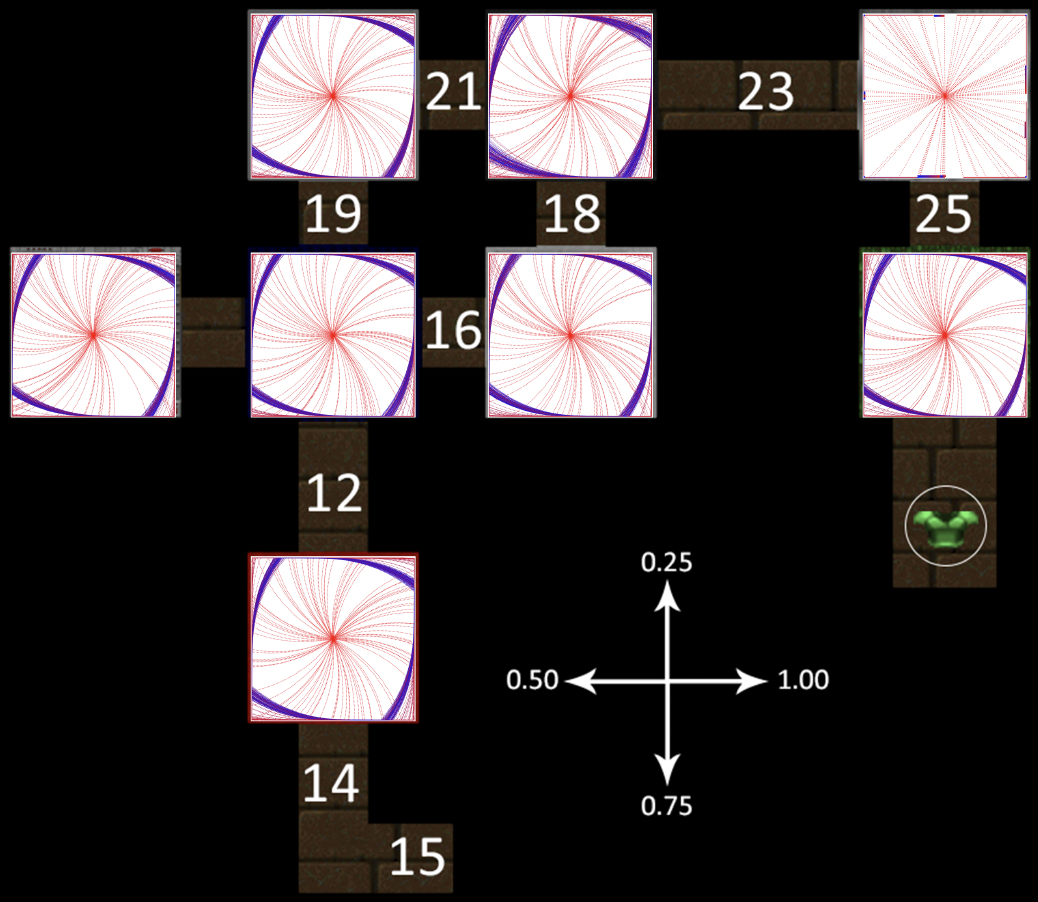}
\caption{TPG paths in the multi-empty room scenario. Each room is parameterized as per Figure \ref{fig:large} but using the wall features corresponding to each of the original labyrinth rooms (Figure \ref{fig:myh_map}). Results superimposed on the original labyrinth layout for easier association. Room 24 is between corridors 23 and 25 at the top RHS}\label{fig:all_rooms}
\end{center}
\end{figure*}

Given this new insight into the impact of exposing an agent to a room without any goal object or exits we can repeat the experiment for the same TPG individual, but this time using room colors/textures that match the original rooms from the labyrinth (Figure \ref{fig:large} assumed a single room with colors/textures from the corridors). Figure \ref{fig:all_rooms} summarizes this experiment by superimposing each single room / no goal state experiment on the original labyrinth room layout. It is now apparent that the room color/texture only has a significant impact on `room 24'. In short, the TPG agent appears to adopt the same process for `reorientation' for all but room 24. Room 24 appears to be a specific case in which the agent instead assumes that it is already entering the room on a particular trajectory/orientation (see for example Figures \ref{fig:TPG_paths}(b) and (c)).\footnote{Room 24 is the top RHS, thus in Figures \ref{fig:TPG_paths}(b) and (c) an agent typically enters room 24 on a specific trajectory from corridor 23.} Given that most agent spawn points initialize the agent in rooms that are `up stream' from room 24, such a policy results in higher cumulative rewards.

In short, from the perspective of Braitenberg behaviours, the TPG champion's strategy amounts to adopting a slow arc with a right hand/clockwise trajectory from any initial state. This trajectory continues until a wall is encountered at which point the agent aligns itself to be parallel with the wall in the direction it encountered the wall. After which the slow right hand trajectory is again selected. Depending on the length remaining in the wall, the agent will manage to arc away from the wall, or will mimic a wall following behaviour. Note, however, that this simplified heuristic does not capture other properties, such as what happens on exiting a labyrinth corridor (where there is no wall to immediately follow) or how to avoid getting stuck in corners. The former appears to be addressed by alternating the direction of the arc (relative to the previous choice). The latter is resolved by either turning 90 degrees and following the next wall, or a 180 degree turn is executed and the agent returns back along the same wall (again with the choice alternating).

\subsection{Complexity of TPG champions}
It is also possible to comment on the complexity of TPG solutions where there is no attempt to `tune' these properties, they are entirely emergent and a function of the behavioural interaction between agent and task. In short, on average there are $3.4 \pm 2.8$ ensembles (nodes) per TPG champion with on average $13.6 \pm 11.2$ learners across the TPG graph. Programs would on average index $36.2 \pm 9.6$ pixels if they were context programs and $27.6 \pm 10.3$ pixels if they were action programs. Given that the original state space is $160 \times 120 = 19,200$ pixels this means that TPG solutions decompose the state space to $\leq 0.2\%$ of the original state space per program or $\leq 0.8\%$ of the state space indexed per decision. Figure \ref{fig:tpg_champ} summarizes the particular structure of the TPG champion discussed in Sections \ref{sec:behaviour} through \ref{sec:no_goal}.

\section{Conclusion}\label{sec:conc}
The ability to evolve navigation strategies from partially observable, high-dimensional state information is investigated with the TPG genetic programming framework parameterized to provide `simple’ solutions. That is to say, we have limited TPG to an instruction set consisting of arithmetic operations alone. There is no support for deploying convolutional operators that could potentially search the visual space for particular objects. This means that the interface through which the resulting solutions experience state is very sparse, i.e. $< 1.0\%$ of the state space. This appears to introduce a bias to discovering simple Braitenberg style heuristics for structuring the agent’s navigation behaviour. Properties of the heuristic include the ability to:

\begin{itemize}
\item seek out a room's wall after spawning in the centre of a room;
\item alternate the direction of a slow arcing trajectory after pursuing a wall following behaviour;
\item reorientate after encountering a room's corner;
\end{itemize}

\noindent Discovering such simple heuristics for navigation is therefore a function of the constraints under which TPG was forced to operate.

Previous work with TPG did not discover such policies as either agents experienced multiple unrelated tasks simultaneously, resulting in complex agents that indexed significant amounts of the state space \cite{smith18}. The one study where TPG agents were evolved against single ViZDoom tasks under the same concise parameterization (\cite{bayer21}) was limited by an error in the simulation environment.\footnote{ViZDoom as distributed under the Windows operating system does not select spawn points with uniform probability (see discussion and solution in \cite{bayer23}).} Research with deep learning frameworks has tended to emphasize discovering shortest paths or navigation from/to arbitrary points as opposed to discovering simple navigation heuristics.

Future work will continue to consider the interaction between task geometry and complexity of the resulting solution heuristics. Thus, if room/corridor sizes are asymmetric or have a different number of sides, what impact does this have on the ability to derive Braitenberg style navigation heuristics? Likewise, if the entry and exit points are no longer centred in the middle of each wall, what impact will this have on the heuristic?

\bibliographystyle{plain}
\bibliography{CoG-2023}

\end{document}